\PassOptionsToPackage{dvipsnames, table}{xcolor}
\documentclass[11pt]{article}
\RequirePackage[dvipsnames, table]{xcolor}  

\usepackage{acl}

\author{Deuksin Kwon$^{1,2}$\hspace{0.3cm}Kaleen Shrestha$^1$\hspace{0.3cm}Bin Han$^{1,2}$\hspace{0.3cm}\\\textbf{Elena Hayoung Lee}$^1$\hspace{0.3cm}\textbf{Gale M. Lucas}$^{1,2}$\\
$^1$University of Southern California
$^2$USC for Institute of Creative Technologies\\
\texttt{\{deuksink, kshresth, binhan\}@usc.edu}\\
\texttt{elena.lee@marshall.usc.edu}
\\\texttt{lucas@ict.usc.edu}}

\usepackage{graphicx}
\usepackage{subcaption} 
\usepackage{threeparttable} 
\usepackage{makecell}
\usepackage{amsmath}
\usepackage{commath}
\usepackage{graphicx}
\usepackage{calc}
\usepackage{footmisc}
\usepackage{amsmath}
\usepackage{booktabs}
\usepackage{siunitx}
\usepackage{enumerate}  
\usepackage{multirow}
\usepackage{enumitem} 
\usepackage{float} 
\usepackage{acl}
\usepackage{placeins}
\usepackage{times}
\usepackage{latexsym}
\usepackage{array}
\usepackage{tcolorbox}
\usepackage{adjustbox}
\usepackage[T1]{fontenc}

\usepackage[table,xcdraw]{xcolor}
\usepackage{colortbl} 
\newenvironment{myexampleblock}[1]{%
    \tcolorbox[%
    noparskip,
    title=#1,
    colback=blue!10,
    colframe=blue!75!black]}%
    {\endtcolorbox}

\usepackage[utf8]{inputenc}


\usepackage{fancyvrb}
\usepackage{caption}
\usepackage{newfloat}
\usepackage{fvextra}

\DeclareFloatingEnvironment{boxes} 

\newtcolorbox{theoremblock}[1][]{colframe=white, colbacktitle=green!80!black,
  colback=green!10, coltitle=white, boxrule=0pt, rounded corners=all, 
  fonttitle=\bfseries, title=#1}

\newtcolorbox{propositionblock}[1][]{colframe=white, colbacktitle=red!80!black,
  colback=red!10, coltitle=white, boxrule=0pt, rounded corners=all, width=\textwidth,
  fonttitle=\bfseries, title=#1}

\newtcolorbox{definitionblock}[1][]{colframe=white, colbacktitle=blue!80!black,
  colback=blue!10, coltitle=white, boxrule=0pt, rounded corners=all, width=\textwidth,
  fonttitle=\bfseries, title=#1}

\newtcolorbox[number within=section]{lemmablock}[1][]{
colframe=white, colbacktitle=blue!50!black,
  colback=blue!5, coltitle=white, boxrule=0pt, rounded corners=all, width=\textwidth,
  fonttitle=\bfseries, title=#1,  left=0mm, right=0mm}

\DefineShortVerb{\|} 

\usepackage{microtype}

\usepackage{inconsolata}

\title{Evaluating Behavioral Alignment in Conflict Dialogue:\\A Multi-Dimensional Comparison of LLM Agents and Humans}

\begin{document}
\maketitle

\begin{abstract}


Large Language Models (LLMs) are increasingly deployed in socially complex, interaction-driven tasks, yet their ability to mirror human behavior in emotionally and strategically complex contexts remains underexplored. This study assesses the behavioral alignment of personality-prompted LLMs in adversarial dispute resolution by simulating multi-turn conflict dialogues that incorporate negotiation. Each LLM is guided by a matched Five-Factor personality profile to control for individual variation and enhance realism. We evaluate alignment across three dimensions: linguistic style, emotional expression (e.g., anger dynamics), and strategic behavior. GPT-4.1 achieves the closest alignment with humans in linguistic style and emotional dynamics, while Claude-3.7-Sonnet best reflects strategic behavior. Nonetheless, substantial alignment gaps persist. Our findings establish a benchmark for alignment between LLMs and humans in socially complex interactions, underscoring both the promise and the limitations of personality conditioning in dialogue modeling.  
\end{abstract}

\section{Introduction}
\label{sec:intro}

Large Language Models (LLMs) are increasingly used to simulate human behavior in socially grounded, interactive tasks. Recent efforts enhance human-likeness by assigning personas or personality traits to LLM agents~\cite{serapiogarcía2023personality, jiang2023evaluating}. However, it remains unclear whether personality-prompted LLMs can replicate the behavioral dynamics of real humans, especially in emotionally charged and strategically complex contexts such as conflict resolution.

Dispute resolution involves negotiating interpersonal conflicts and requires emotional expression, strategic adaptation, and relationship management~\cite{brett2007negotiating}. Unlike cooperative bargaining, it includes blame, justification, and tension regulation. While LLMs have been studied in negotiation settings~\cite{abdelnabi2023llm, bianchi2024well}, few works assess whether their behavior aligns with humans in adversarial, emotionally charged interactions.

\begin{figure}
    \centering
    \includegraphics[width=\columnwidth]{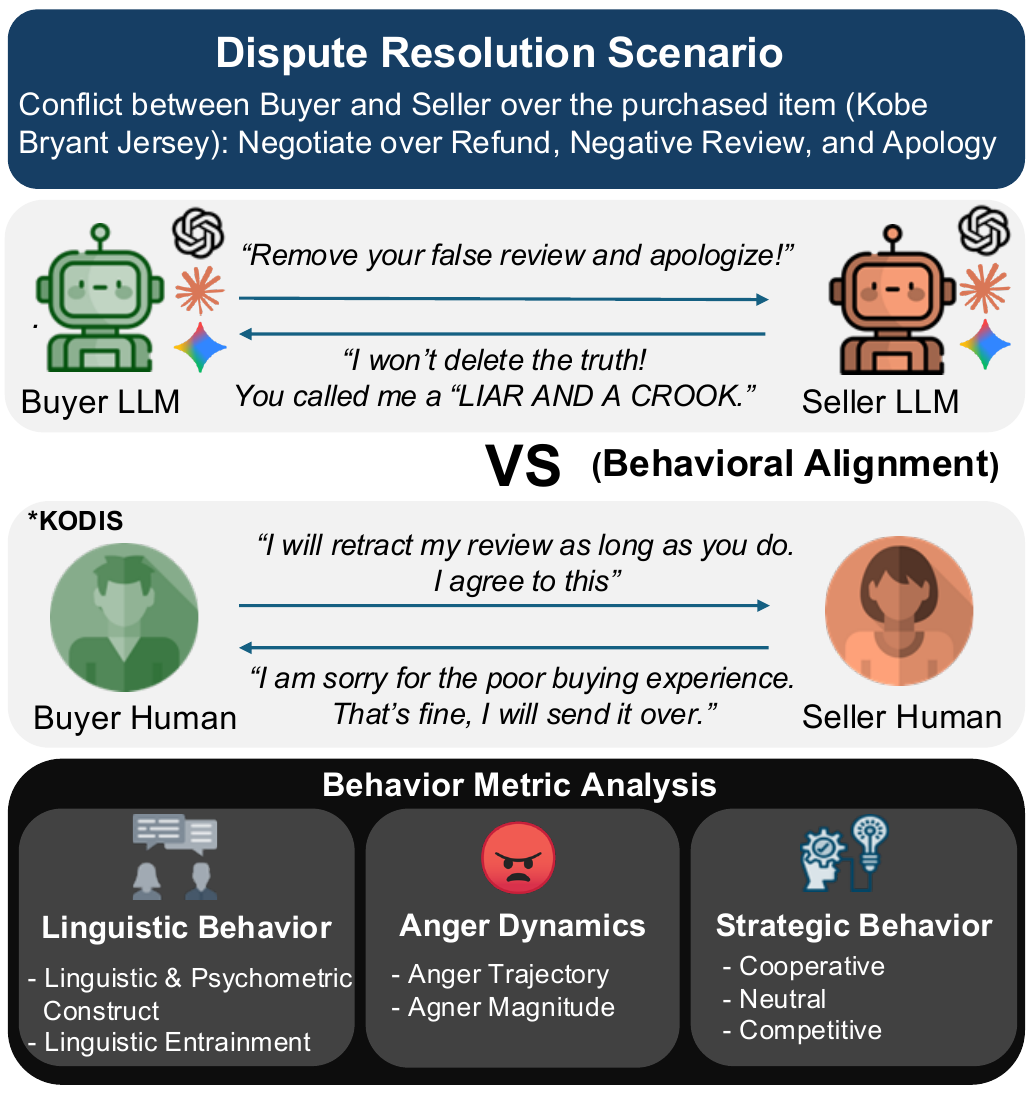}
    \caption{Overview of a dispute resolution scenario examining the behavioral alignment between LLM agents and humans, analyzed across linguistic behavior, anger dynamics, and strategic behavior.}
    \label{fig:intro}
\end{figure}

We address this gap through a comparative study using multi-turn, multi-issue human negotiation dialogues~\cite{hale2025kodis}, simulating LLM agents prompted with matched Five-Factor Model personality profiles~\cite{mccrae1992introduction, abdurahman2024perils}. To enable systematic comparison, we introduce a structured evaluation framework with tailored metrics that capture alignment across three behavioral dimensions: (1) Linguistic Style (LIWC, nClid)~\cite{ireland2014language}, (2) Emotional Dynamics (turn-level anger trajectories), and (3) Strategic Behavior (\textit{Interests-Rights-Power} (IRP) transitions)~\cite{ury1988getting}. These metrics enable rigorous, multi-dimensional comparisons that reveal both surface-level patterns and deeper behavioral differences between human and agent interactions. Figure~\ref{fig:intro} provides an overview of our experimental framework and the dispute resolution scenario.

Results show that GPT-4.1 aligns most closely with humans in language and emotional expression, while Claude-3.7-Sonnet better reflects strategic negotiation behavior. Despite this progress, both models display meaningful deviations, highlighting the limits of current LLMs in modeling socially and emotionally nuanced interaction. Our framework offers a replicable basis for benchmarking LLM-human alignment in complex dialogue. Our main contributions are as follows:
\begin{itemize}
    \item We present a structured, behaviorally grounded evaluation of personality-prompted LLMs in adversarial negotiation, using a multi-perspective framework to compare their linguistic, emotional, and strategic behaviors with human data.
    \item We identify both the strengths and systematic behavioral gaps of state-of-the-art LLMs, including GPT-4.1 and Claude-3.7-Sonnet. 
    \item We offer critical insight into the fidelity of LLMs’ social behavior under personality conditioning, highlighting its promise and limitations in agent-based simulations.
\end{itemize}

\section{Related Works}
\label{sec:related work}

LLMs have been increasingly used for agent-based simulations, showing promise in role-playing, personality conditioning, and goal-oriented dialogue~\cite{li-etal-2024-econagent, kwon2025astra, gao2024large, gloria-silva-etal-2024-plan, jiang2022mpi}. Prior work has examined whether LLMs can emulate human traits such as trust, personality, or emotion-driven behavior, but mainly in single-turn settings~\cite{wang-etal-2024-rolellm, wang2025evaluating, xie2024can}. These studies often lack multi-turn interaction and overlook the evolving nature of behavior~\cite{wang-etal-2024-rolellm}. Even in dyadic setups, comparisons typically focus on outcomes rather than on coordination during the dialogue~\cite{xie2024can}.

Despite the central role of negotiation in human communication, prior work has not directly compared LLM and human dyads in extended, emotionally complex interactions~\cite{gandhi2023strategic, noh2024llms, huang-hadfi-2024-personality}. This gap is especially salient in dispute resolution, where negotiation occurs amid pre-existing conflict and heightened emotions such as anger. Unlike cooperative settings, dispute resolution requires both strategic reasoning and emotional sensitivity~\cite{brett2007negotiating}.

To fill this gap, we present the first systematic comparison of LLMs and humans in multi-turn, adversarial dialogue. Using the socially and emotionally rich context of dispute resolution, we evaluate agent-human alignment across three key dimensions: linguistic style, emotional dynamics, and strategic behavior. 

\section{Methods}

\subsection{Human Dataset}
The \emph{KObe DISpute corpus} (KODIS) is a human–human (H2H) dispute resolution dataset consisting of extended English role-play dialogues between individuals assigned to Buyer and Seller roles~\cite{hale2025kodis}. The scenario centers on a dispute over a jersey purchased online for the Buyer’s sick nephew, with both parties expressing conflicting perspectives and strong negative emotions. Participants negotiate a resolution across three predefined issues: (1) a full refund, (2) removing a negative review, and (3) offering a formal apology. We use the KODIS dataset for its naturalistic human conflict dialogues with rich emotional and strategic annotations, making it well-suited for examining behavioral alignment with LLM dialogues. For our study, we exclude human–AI dialogues and focus on 248 H2H conversations with complete personality data for both participants. All analyses are conducted on filtered subsets with missing values removed as needed.

\subsection{LLM-LLM (L2L) Simulated Dialogue}
To evaluate how LLMs behave in conflict resolution, we simulate dyadic negotiations between two LLMs (Buyer and Seller) using the KODIS scenario. Each LLM is assigned a distinct personality profile based on the Big Five Inventory, structured across six levels that reflect both trait polarity (positive vs. negative) and degree (low, medium, high). These traits are verbalized using personality-relevant adjectives, following the approach of~\citet{huang-hadfi-2024-personality}. In addition, each LLM is given a personalized issue importance profile and negotiates more assertively on issues it values more. To ensure fair comparison with human data, prompts mirror those in KODIS, and agent traits and priorities are weighted-sampled to match human distributions. (see Appendix Figure~\ref{append:personality_profiling} for personality distribution comparison). Dialogues span five negotiable issues (e.g., refund, apology, review) and proceed turn by turn until one agent accepts an offer or walks away. 

We simulate 250 dispute resolution dialogues for each of four widely used LLMs: GPT-4.1-mini and GPT-4.1 (OpenAI), Claude-3.7-Sonnet (Anthropic, hereafter Claude), and Gemini-2.0-Flash (Google, hereafter Gemini). Information on the decoding hyperparameters can be found in Appendix~\ref{append:hyperperameter}, and further details on prompt design and personality construction are provided in Appendix~\ref{append:personality_setting}. Simulation code will be made publicly available\footnote{\url{https://github.com/DSincerity/Eval-LLM-BehavAlign}}.



\subsection{Behavior Metrics}
\subsubsection{Linguistic and Psychometric Construct Gap using LIWC (LG)}~\label{sec:LG}
To quantify the usage of certain words pertaining to linguistic and psychometric constructs related dispute resolution, we extract a set of 10 Linguistic Inquiry and Word Count (LIWC) \cite{boyd2022development} features from the entire corpus of conversations for each dataset. We compare the LIWC feature gap (LG) between conversations from human participants in the KODIS dataset and each LLM with the Jensen-Shannon Divergence (JSD)~\cite{fuglede2004jensen}\footnote{We used the SciPy python package v1.13.1 for JSD.} used to measure the divergence between two probability distributions and ranges from 0 (identical) to 1 (maximally different):

\begin{equation}
\small
\text{LG}_{\text{LLM}} = \abs{\overline{\text{JSD}}_{\text{Within-Human}} -\overline{\text{JSD}}_{\text{LLM-Human}}}
\end{equation}

\noindent where $\overline{\text{JSD}}_{Within-Human}$ denotes the average JSD of the LIWC features distribution between two conversations in KODIS, and $\overline{\text{JSD}}_{LLM-Human}$ denotes the average JSD of LIWC features distribution between two conversations, one in the LLM dataset, the other in KODIS. Lower $\text{LG}_{LLM}$ values indicate that LLM uses language related to the selected LIWC categories similar to that of humans. The 10 LIWC constructs we selected were split into two categories: (1) dispute resolution strategies, specifically interests-rights-power (IRP) strategies ~\cite{brett2007negotiating} (LG - IRP: \textit{insight, prosocial behavior, affiliation, power, all-or-none, and politeness}) and (2) the KODIS scenario or dispute in general (LG - Dispute: \textit{money, analytical thinking, authentic, clout}). 

\subsubsection{Linguistic Entrainment Gap (LEG)}
Effective conflict resolution often involves linguistic entrainment (LE), where speakers adapt their language to align with their partner~\cite{taylor2008linguistic}. To assess whether LLM dyads exhibit human-like LE, we use the normalized conversational linguistic distance (nCLiD)~\cite{Nasir2019ModelingIL}, which measures conversation-level linguistic coordination. As nCLiD is directional, we compute LE for both directions (buyer to seller and vice versa) and average the two to obtain a dyadic LE score. The formula for nCLiD is provided in the Appendix~\ref{append:LE}. We compute the difference in average LE scores between human and LLM conversations:

\vspace{-2ex}
\begin{equation}
\text{LEG}_{\text{LLM}} = \abs{  \overline{\text{LE}}_{\text{LLM}} -  \overline{\text{LE}}_{\text{Human}} }
\end{equation}

\noindent where $\overline{\text{LE}}_{\text{LLM}}$ and $\overline{\text{LE}}_{\text{Human}}$ are the average of LE scores for all dyads of H2H and L2L conversations, respectively. Lower LEG indicates closer alignment of LLM linguistic entrainment with humans.

\begin{table*}[th!]
\centering
\resizebox{\textwidth}{!}{%
\begin{threeparttable}
\begin{tabular}{lcccccc}
\hline
\multicolumn{1}{c}{\multirow{3}{*}{Model}} & \multicolumn{6}{c}{Metrics}   \\ \cline{2-7} 
\multicolumn{1}{c}{} & \multicolumn{3}{c}{Linguistic Features} & \multicolumn{2}{c}{Anger Dynamics} & Strategic Behavior \\ \cline{2-7} 
\multicolumn{1}{c}{} & LG - IRP\tnote{‡} & LG - Dispute\tnote{‡} & LEG\tnote{†} & ATG\tnote{‡} & AMG\tnote{†} & SBG\tnote{‡} \\ \hline
GPT4.1         & 0.041$^{***}$ & \textbf{0.021$^{***}$} & \textbf{0.004}$^{**}$ & \textbf{0.195}$^{***}$ & \textbf{0.183}$^{***}$ & 0.103$^{***}$ \\
GPT4.1-mini    & 0.040$^{***}$ & 0.043$^{***}$ & 0.011$^{***}$ & 0.465$^{***}$ & 0.186$^{***}$ & 0.125$^{***}$ \\
Gemini  & \textbf{0.036}$^{***}$ & 0.053$^{***}$ & 0.014$^{***}$ & 0.345$^{***}$ & 0.212$^{***}$ & 0.102$^{***}$ \\
Claude & 0.046$^{***}$ & \textbf{0.021}$^{***}$ & 0.007$^{***}$ & 0.363$^{***}$ & 0.367$^{***}$ & \textbf{0.018}$^{***}$ \\ \hline
\begin{tabular}[c]{@{}l@{}}KODIS \\ (Human Baseline)\end{tabular} &
\begin{tabular}[c]{@{}c@{}}Avg. Within-Human\\ JSD: 0.179\end{tabular} &
\begin{tabular}[c]{@{}c@{}}Avg. Within-Human\\ JSD: 0.128\end{tabular} &
\begin{tabular}[c]{@{}c@{}}Avg. Human nCliD \\ Score: 0.311\end{tabular} &
\begin{tabular}[c]{@{}c@{}}Avg. Within-Human \\ DTW distance: 0.86\end{tabular} &
\begin{tabular}[c]{@{}c@{}}Avg. AUC of Human Anger \\ Trajectory: 0.286\end{tabular} &
\begin{tabular}[c]{@{}c@{}}Avg. Within-Human \\ JSD: 0.127\end{tabular} \\ \hline
\end{tabular}
\begin{tablenotes}
\large
\item[†] Dyad-level metrics: averaged across dyads (e.g., LEG, AMG).
\item[‡] Distribution-level metrics (“Within-Human”): computed from pairwise comparisons over the full result distribution (e.g., LG, ATG, SBG). \\
Statistical significance : * p < .05, ** p < .01, *** p < .001 (independent t-test vs. human baseline).
\end{tablenotes}
\end{threeparttable}
}
\caption{Divergence metric results for all LLMs compared to KODIS, along with the baseline for KODIS.}
\label{tab:metric_results}
\end{table*}


\subsubsection{Emotion (Anger) Dynamics}

\noindent \textbf{Anger Trajectory Gap (ATG)}
We compare anger dynamics between H2H and L2L dyads using two complementary metrics, as anger is an influential factor in shaping communication and outcomes in dispute resolution. Anger intensity was annotated using a pretrained BERT-based classifier\footnote{https://huggingface.co/tae898/emoberta-large}~\cite{kim2108emoberta}. First, to assess how similarly anger unfolds over time, we use Dynamic Time Warping (DTW), a method widely used to compare time series as it allows non-linear alignment and handles sequences of different lengths~\cite{muller2007dynamic}.

To quantify how similar each LLM’s anger dynamics are to those of humans, we define ATG using DTW as follows:
\vspace{-2ex}

\begin{equation}
\small
\text{ATG}_{\text{LLM}} = \abs{\overline{\text{DTW}}_{\text{Within-Human}} -\overline{\text{DTW}}_{\text{LLM-Human}}}
\end{equation}


\noindent where \( \overline{\text{DTW}}_{\text{Within-Human}} \) denotes the average DTW distance between anger intensity trajectories in human-human pairs, and \( \overline{\text{DTW}}_{\text{LLM-Human}} \) denotes the average DTW distance between trajectories in LLM-human pairs. A lower \( \text{ATG}_{\text{LLM}} \) indicates that the LLM's anger dynamics more closely resemble those observed in human interactions.

\noindent \textbf{Anger Magnitude Gap (AMG)}
Second, to compare the overall magnitude of anger expressed during interactions, we compute the difference in area under the anger intensity curves (AUC):

\vspace{-2ex}
\begin{equation}
\text{AMG}_{\text{LLM}} = \abs{  \overline{\text{AUC}}_{\text{LLM}} -  \overline{\text{AUC}}_{\text{Human}} }
\end{equation}

\noindent Here, $\overline{\text{AUC}}_{\text{Human}}$ denotes the average AUC of anger intensity trajectories from all H2H dyads, while $\overline{\text{AUC}}_{\text{LLM}}$ denotes the average across all L2L dyads. The resulting difference quantifies the gap in overall anger magnitude, reflecting how closely the LLM’s emotional intensity aligns with humans. This dual-metric approach captures both the trajectory and intensity of anger, offering a comprehensive view of emotional alignment.


\subsubsection{Strategic Behavior Gap (SBG)}
To compare strategic behavior, we measure how similarly agents and humans distribute their use of IRP strategies, which capture how people strategically navigate conflict~\cite{ury1988getting}. We compute distributional gaps using the JSD metric described in Section ~\ref{sec:LG}:
\vspace{-2ex}

\begin{equation}
\small
\text{SBG}_{\text{LLM}} = \abs{\overline{\text{JSD}}_{\text{Within-Human}} -\overline{\text{JSD}}_{\text{LLM-Human}}}
\end{equation}

\noindent where \( \overline{\text{JSD}}_{\text{Within-Human}} \) denotes the average JSD of IRP usage distribution in human-human pairs, and \( \overline{\text{JSD}}_{\text{LLM-Human}} \) denotes the average JSD of IRP usage distribution in all LLM-human pairs.

A lower \( \text{SBG}_{\text{LLM}} \) indicates that LLM uses a strategic profile more similar to that of humans.





\section{Results and Analysis}

\subsection{Key Behavioral Outcomes}

Table~\ref{tab:descriptive-stats} in Appendix summarizes outcomes from simulated dispute conversations. LLMs tended to negotiate longer than humans, with Gemini averaging the most rounds (11.63) and Claude the fewest. Walk-away rates ranged widely, from Gemini’s high (0.53) to Claude’s near-zero, indicating stronger resolution efforts. Claude’s deal score gaps closely matched humans, while Gemini’s low gap (13.51) may suggest overly balanced or less satisfying outcomes.

\subsection{Linguistic and Psychometric Construct
Gap using LIWC (LG)}
Comparing the LG scores of the four LLMs (Table \ref{tab:metric_results}) for IRP-related LIWC categories, Gemini has the lowest divergences from KODIS with an LG of 0.036, while Claude had the highest difference from human conversations. For the LG scores related to dispute in general, both GPT and Claude had the lowest difference from KODIS with an LG score of 0.021, while Gemini, interestingly, had the highest difference. This suggests that Gemini models generate more human-like language related to IRP categories, but not as much for LIWC categories related to dispute in general. Similarly, Claude and GPT-4.1 apparently generate more human-like language related to dispute in general, but not as much for IRP-related words.

\subsection{Linguistic Entrainment Gap (LEG)}
For the gap between linguistic entrainment between the buyer and seller in human conversations and LLM conversations (Table \ref{tab:metric_results}), we see that GPT-4.1 again has the smallest difference of linguistic entrainment found in the KODIS dataset with a LEG score of 0.004. Here we find that Gemini again has the highest difference in LE score with a LEG score of 0.0014. This suggests that GPT-4.1 has the most similar linguistic entrainment between buyers and sellers as found in human conversations in the KODIS dataset.

\subsection{Anger Dynamics}
\begin{figure}[t]
  \centering
  \begin{subfigure}{\linewidth}
    \centering
    \includegraphics[width=0.8\linewidth]{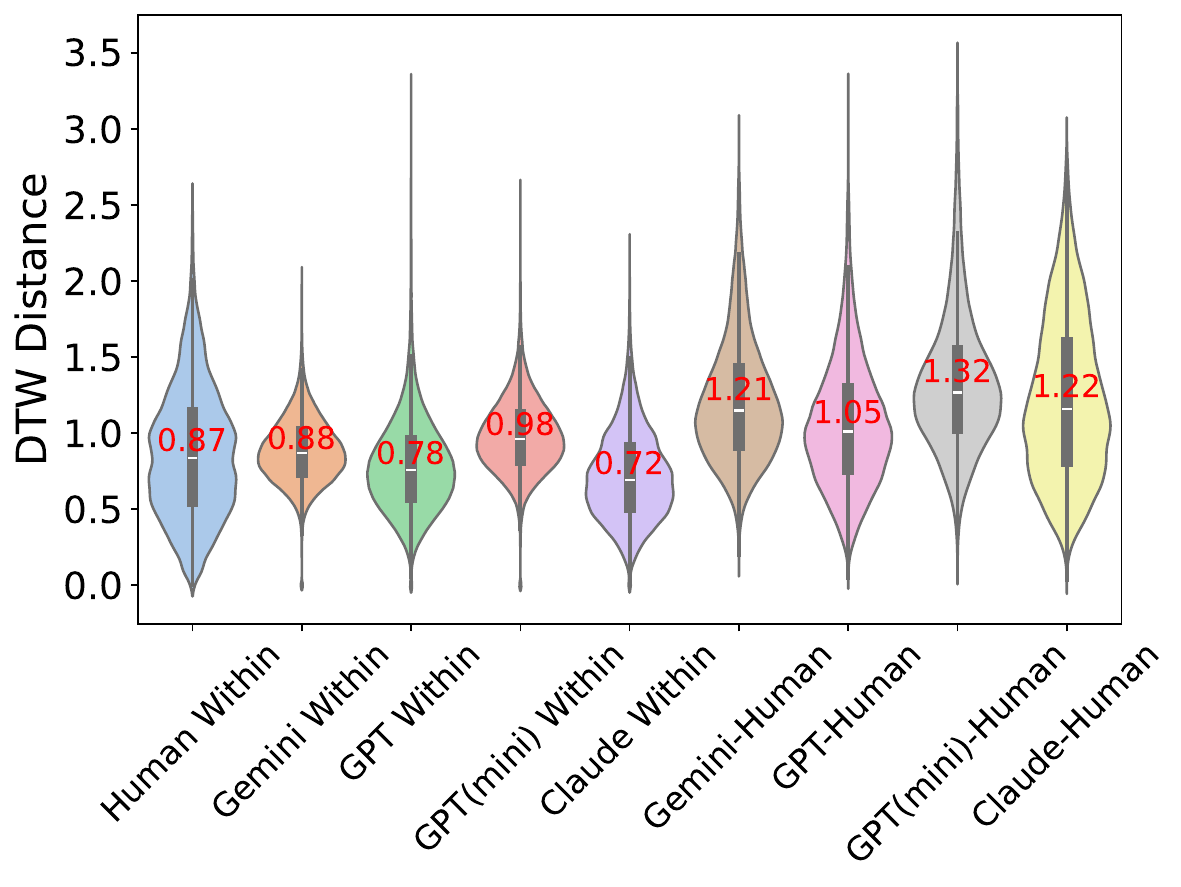}
    \caption{Distribution of DTW distances for Humans and LLMs ('Within' = comparisons within the same group; 'LLM–Human' = pairwise comparisons across groups)}
    \label{fig:DTW}
  \end{subfigure}

  \vspace{1em} 

  \begin{subfigure}{0.95\linewidth}
    \centering
    \includegraphics[width=0.8\linewidth]{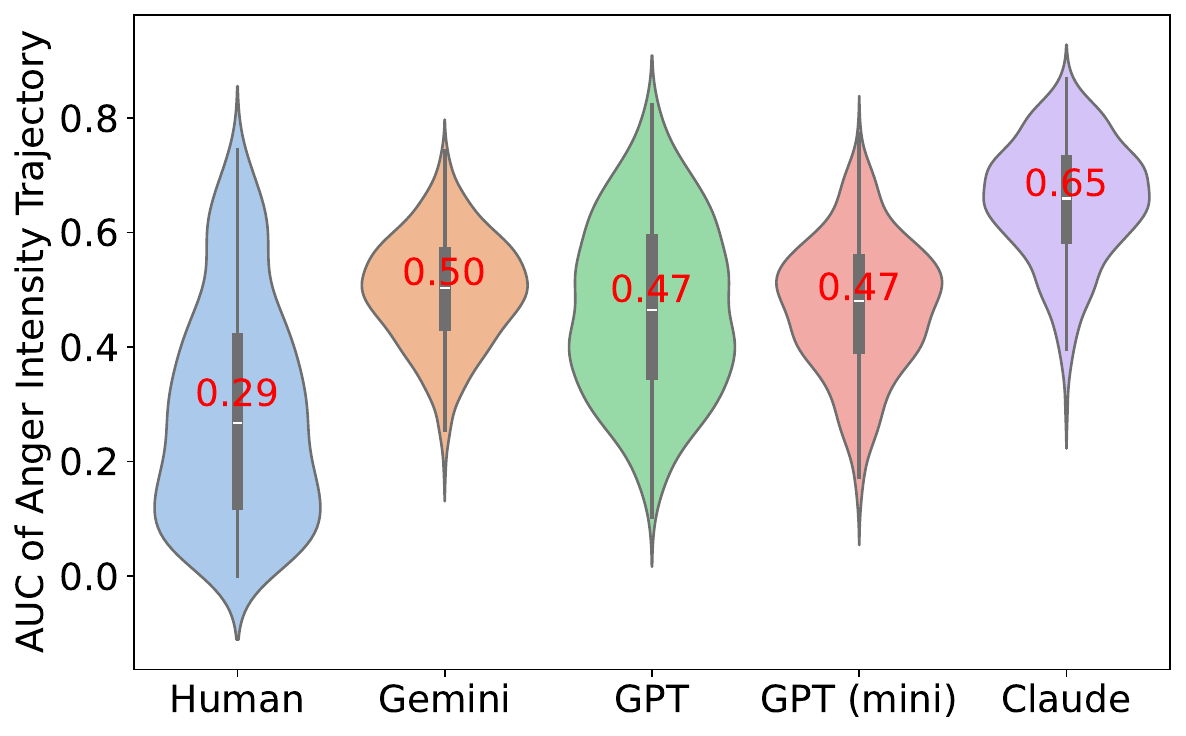}
    \caption{Distribution of Anger Intensity Trajectory AUCs for Humans and LLMs}
    \label{fig:AUCs}
  \end{subfigure}
  \caption{Comparison of human and LLM distributions for (a) DTW distances and (b) anger intensity AUCs. \\ {\footnotesize * Models: GPT(-mini) = GPT-4.1(-mini), Gemini = Gemini-Flash-2.0, Claude = Claude-3.7-Sonnet}}
  \label{fig:dtw_auc_combined}
\end{figure}

Based on the DTW distance between agents and humans (Table~\ref{tab:metric_results}), GPT-4.1 exhibits the smallest ATG at 1.05, while GPT-4.1-mini shows the largest. Given that the average within-human DTW similarity in the KODIS dataset is around 0.86, GPT-4.1 approximates human anger dynamics relatively closely. However, as shown in Figure~\ref{fig:DTW}, the within-model DTW distributions (LLM-within) reveal a key difference in variability: human dyads show high variance in anger trajectories, while LLM dyads exhibit lower variance overall. Among them, GPT-4.1 and Claude stand out with relatively higher variance, more closely mirroring the diversity seen in human interactions. In the cross-model DTW distribution (LLM–Human), GPT-4.1 again shows the lowest mean and relatively low variance, confirming its closer alignment with human anger trajectories as shown in Table~\ref{tab:metric_results}.

In terms of anger magnitude, agreeable LLMs surprisingly express higher levels of anger in dispute resolution scenarios than humans (mean AUC for humans: 0.286). Although the differences among models are relatively small, Claude shows substantially greater anger expression, whereas GPT-4.1 demonstrates the most human-like magnitude, though the gap from humans remains significant (independent t-test: t = –16.43, p < 0.001). Nonetheless, as shown in Figure~\ref{fig:AUCs}, human anger intensity exhibits broader variance and greater variability; while LLMs generally show reduced variance, GPT-4.1 stands out with relatively higher variance, aligning with humans not only in magnitude but also in diversity.


\subsection{Strategic Behavior Gap (SBG)}
In comparing the use of various IRP strategies for dispute resolution between artificial agents and humans, we found that among LLMs, Claude exhibited the most human-like strategic behavior, as evidenced by a minimal difference in the average within-human JSD of IRP usage distributions (SBG=0.018). As shown in Figure~\ref{fig:SB} of the Appendix, Claude’s IRP distribution closely aligns with the proportions observed in the human KODIS dataset, including its use of strategies such as Fact and Power. This alignment suggests a strong convergence between Claude’s decision-making and human strategic reasoning in interpersonal conflict scenarios.

Interestingly, as with humans, linguistic and strategic behavior do not always align: Claude’s lower LG-IRP score in Table~\ref{tab:metric_results} suggests its language use is less tied to IRP markers, yet its strong SBG-IRP performance shows close alignment with human strategies. This highlights that IRP strategies may not be fully captured by surface-level word choice, underscoring the distinction between linguistic cues and underlying strategic behavior.
\section{Conclusion}

We present divergence metrics to compare linguistic and behavioral characteristics of LLM and human dispute resolution conversations, focusing on linguistic entrainment (LEG), language use in IRP strategies and dispute contexts (LG), anger dynamics, and strategic behavior (SBG). Our results show that GPT-4.1 exhibits the closest behavioral alignment with humans across most metrics, though notable gaps remain in strategic behavior and IRP-related language use. These findings highlight both the current progress and the limitations of LLMs in socially complex interactions, suggesting future directions for improving alignment through more nuanced modeling of strategic reasoning and emotional dynamics.

\clearpage
\section*{Limitations}
Due to resource limitations, we were unable to test a wider range of LLMs, and in the future would like to rank open-source models to get a fuller picture of the range of performance of LLMs in comparison to humans. Another limitation of this work is that KODIS dataset scenario is not realisitic negotiation conversations, since it is a role-play setup, and thus may not relect true dispute/negotiation human behavior. In some cases, LLMs displayed inconsistencies between their stated issue importance and their negotiation behavior, suggesting a lack of strategic reasoning. Developing improved prompting strategies or alternative methods to better support LLMs’ strategic decision-making could lead to more realistic simulations and enable more accurate comparisons with human behavioral patterns. 
 
\section*{Ethical Considerations}
\subsection*{Dataset}
Our study uses the KODIS dataset~\cite{hale2025kodis}, a publicly available corpus of dispute resolution dialogues that was collected through crowdsourcing and has been fully anonymized. We strictly adhered to the dataset’s licensing terms, intended use, and ethical guidelines. All dialogues are in English, and all simulations for both the baseline and our proposed agent were conducted exclusively in English. The dataset contains no personally identifiable information, and our use was solely for academic research purposes.

\subsection*{LLMs}
We used LLMs strictly in accordance with their intended functions and licensing agreements, ensuring alignment with ethical norms and regulatory requirements. Consistent with recent research utilizing LLM-based agent simulations, our methodology promotes transparency and responsibility while adhering to established usage guidelines.

\subsection*{Use AI assistant Tools}
We leveraged AI tools such as ChatGPT to help with language polishing and to support code debugging and enhancement. Nonetheless, all key ideas, experimental setups, algorithmic designs, methodologies, and final implementations were solely conceived, executed, and verified by the authors.

\subsection*{Potential Risks}
Using LLMs to simulate human behavior is an active area of research, as LLMs are known to simplify or otherwise inaccurately represent human behaviors \cite{abdurahman2024perils}, as also seen in our findings with the divergence of the LLM agents from human conflict resolution behavior. Additionally, any use of LLMs, especially in the context of conflict, should be carefully studied before deployment in real human-facing applications due to emotional reactions conflict can evoke. 

\section*{Acknowledgements}
Research was sponsored by the Army Research Office under Cooperative Agreement Number W911NF-25-2-0040. The views and conclusions contained in this document are those of the authors and should not be interpreted as representing the official policies, either expressed or implied, of the Army Research Office or the U.S. Government. The U.S. Government is authorized to reproduce and distribute reprints for Government purposes, notwithstanding any copyright notation herein. 

Kaleen Shrestha is supported in part by the NSF CISE Graduate Fellowship CSGrad4US under Grant No. 2313998 (Award ID G-2A-061).

\clearpage
\bibliography{anthology,custom}

\begin{thebibliography}{33}
\expandafter\ifx\csname natexlab\endcsname\relax\def\natexlab#1{#1}\fi

\bibitem[{Abdelnabi et~al.(2023)Abdelnabi, Gomaa, Sivaprasad, Sch{\"o}nherr, and Fritz}]{abdelnabi2023llm}
Sahar Abdelnabi, Amr Gomaa, Sarath Sivaprasad, Lea Sch{\"o}nherr, and Mario Fritz. 2023.
\newblock Llm-deliberation: Evaluating llms with interactive multi-agent negotiation games.

\bibitem[{Abdurahman et~al.(2024)Abdurahman, Atari, Karimi-Malekabadi, Xue, Trager, Park, Golazizian, Omrani, and Dehghani}]{abdurahman2024perils}
Suhaib Abdurahman, Mohammad Atari, Farzan Karimi-Malekabadi, Mona~J Xue, Jackson Trager, Peter~S Park, Preni Golazizian, Ali Omrani, and Morteza Dehghani. 2024.
\newblock Perils and opportunities in using large language models in psychological research.
\newblock \emph{PNAS nexus}, 3(7):pgae245.

\bibitem[{Anthony~Hale et~al.(2025)Anthony~Hale, Rakshit, Chawla, M~Brett, and Gratch}]{hale2025kodis}
James Anthony~Hale, Sushrita Rakshit, Kushal Chawla, Jeanne M~Brett, and Jonathan Gratch. 2025.
\newblock Kodis: A multicultural dispute resolution dialogue corpus.
\newblock In \emph{Proceedings of the 2025 Conference of the Nations of the Americas Chapter of the Association for Computational Linguistics: Human Language Technologies (Volume 1: Long Papers)}, page 12771–12785, Albuquerque, New Mexico. Association for Computational Linguistics.

\bibitem[{Bianchi et~al.(2024)Bianchi, Chia, Yuksekgonul, Tagliabue, Jurafsky, and Zou}]{bianchi2024well}
Federico Bianchi, Patrick~John Chia, Mert Yuksekgonul, Jacopo Tagliabue, Dan Jurafsky, and James Zou. 2024.
\newblock How well can llms negotiate? negotiationarena platform and analysis.
\newblock \emph{arXiv preprint arXiv:2402.05863}.

\bibitem[{Boyd et~al.(2022)Boyd, Ashokkumar, Seraj, and Pennebaker}]{boyd2022development}
Ryan~L Boyd, Ashwini Ashokkumar, Sarah Seraj, and James~W Pennebaker. 2022.
\newblock The development and psychometric properties of liwc-22.
\newblock \emph{Austin, TX: University of Texas at Austin}, 10:1--47.

\bibitem[{Brett(2007)}]{brett2007negotiating}
Jeanne~M Brett. 2007.
\newblock \emph{Negotiating globally: How to negotiate deals, resolve disputes, and make decisions across cultural boundaries}.
\newblock John Wiley \& Sons.

\bibitem[{Brett et~al.(1998)Brett, Shapiro, and Lytle}]{brett1998breaking}
Jeanne~M Brett, Debra~L Shapiro, and Anne~L Lytle. 1998.
\newblock Breaking the bonds of reciprocity in negotiations.
\newblock \emph{Academy of Management Journal}, 41(4):410--424.

\bibitem[{Fuglede and Topsoe(2004)}]{fuglede2004jensen}
Bent Fuglede and Flemming Topsoe. 2004.
\newblock Jensen-shannon divergence and hilbert space embedding.
\newblock In \emph{International symposium onInformation theory, 2004. ISIT 2004. Proceedings.}, page~31. IEEE.

\bibitem[{Gandhi et~al.(2023)Gandhi, Sadigh, and Goodman}]{gandhi2023strategic}
Kanishk Gandhi, Dorsa Sadigh, and Noah~D Goodman. 2023.
\newblock Strategic reasoning with language models.
\newblock \emph{arXiv preprint arXiv:2305.19165}.

\bibitem[{Gao et~al.(2024)Gao, Lan, Li, Yuan, Ding, Zhou, Xu, and Li}]{gao2024large}
Chen Gao, Xiaochong Lan, Nian Li, Yuan Yuan, Jingtao Ding, Zhilun Zhou, Fengli Xu, and Yong Li. 2024.
\newblock Large language models empowered agent-based modeling and simulation: A survey and perspectives.
\newblock \emph{Humanities and Social Sciences Communications}, 11(1):1--24.

\bibitem[{Gautam(2014)}]{gautam2014kappa}
Shiva Gautam. 2014.
\newblock A-kappa: a measure of agreement among multiple raters.
\newblock \emph{Journal of Data Science}, 12:697--716.

\bibitem[{Gl{\'o}ria-Silva et~al.(2024)Gl{\'o}ria-Silva, Ferreira, Tavares, Semedo, and Magalhaes}]{gloria-silva-etal-2024-plan}
Diogo Gl{\'o}ria-Silva, Rafael Ferreira, Diogo Tavares, David Semedo, and Joao Magalhaes. 2024.
\newblock \href {https://aclanthology.org/2024.eacl-long.77/} {Plan-grounded large language models for dual goal conversational settings}.
\newblock In \emph{Proceedings of the 18th Conference of the European Chapter of the Association for Computational Linguistics (Volume 1: Long Papers)}, pages 1271--1292, St. Julian{'}s, Malta. Association for Computational Linguistics.

\bibitem[{Goldberg(1992)}]{goldberg1992development}
Lewis~R Goldberg. 1992.
\newblock The development of markers for the big-five factor structure.
\newblock \emph{Psychological assessment}, 4(1):26.

\bibitem[{Huang and Hadfi(2024)}]{huang-hadfi-2024-personality}
Yin~Jou Huang and Rafik Hadfi. 2024.
\newblock \href {https://doi.org/10.18653/v1/2024.findings-emnlp.605} {How personality traits influence negotiation outcomes? a simulation based on large language models}.
\newblock In \emph{Findings of the Association for Computational Linguistics: EMNLP 2024}, pages 10336--10351, Miami, Florida, USA. Association for Computational Linguistics.

\bibitem[{Ireland and Henderson(2014)}]{ireland2014language}
Molly~E Ireland and Marlone~D Henderson. 2014.
\newblock Language style matching, engagement, and impasse in negotiations.
\newblock \emph{Negotiation and conflict management research}, 7(1):1--16.

\bibitem[{Jiang et~al.(2022)Jiang, Xu, Zhu, Han, Zhang, and Zhu}]{jiang2022mpi}
Guangyuan Jiang, Manjie Xu, Song-Chun Zhu, Wenjuan Han, Chi Zhang, and Yixin Zhu. 2022.
\newblock Mpi: Evaluating and inducing personality in pre-trained language models.
\newblock \emph{arXiv preprint arXiv:2206.07550}.

\bibitem[{Jiang et~al.(2023)Jiang, Xu, Zhu, Han, Zhang, and Zhu}]{jiang2023evaluating}
Guangyuan Jiang, Manjie Xu, Song-Chun Zhu, Wenjuan Han, Chi Zhang, and Yixin Zhu. 2023.
\newblock \href {http://arxiv.org/abs/2206.07550} {Evaluating and inducing personality in pre-trained language models}.

\bibitem[{Kim and Vossen(2021)}]{kim2108emoberta}
T~Kim and P~Vossen. 2021.
\newblock Emoberta: Speaker-aware emotion recognition in conversation with roberta. arxiv 2021.
\newblock \emph{arXiv preprint arXiv:2108.12009}.

\bibitem[{Kusner et~al.(2015)Kusner, Sun, Kolkin, and Weinberger}]{Kusner2015FromWE}
Matt~J. Kusner, Yu~Sun, Nicholas~I. Kolkin, and Kilian~Q. Weinberger. 2015.
\newblock \href {https://api.semanticscholar.org/CorpusID:14674248} {From word embeddings to document distances}.
\newblock In \emph{International Conference on Machine Learning}.

\bibitem[{Kwon et~al.(2025)Kwon, Hae, Clift, Shamsoddini, Gratch, and Lucas}]{kwon2025astra}
Deuksin Kwon, Jiwon Hae, Emma Clift, Daniel Shamsoddini, Jonathan Gratch, and Gale~M Lucas. 2025.
\newblock Astra: A negotiation agent with adaptive and strategic reasoning through action in dynamic offer optimization.
\newblock \emph{arXiv preprint arXiv:2503.07129}.

\bibitem[{Li et~al.(2024)Li, Gao, Li, Li, and Liao}]{li-etal-2024-econagent}
Nian Li, Chen Gao, Mingyu Li, Yong Li, and Qingmin Liao. 2024.
\newblock \href {https://doi.org/10.18653/v1/2024.acl-long.829} {{E}con{A}gent: Large language model-empowered agents for simulating macroeconomic activities}.
\newblock In \emph{Proceedings of the 62nd Annual Meeting of the Association for Computational Linguistics (Volume 1: Long Papers)}, pages 15523--15536, Bangkok, Thailand. Association for Computational Linguistics.

\bibitem[{McCrae and John(1992)}]{mccrae1992introduction}
Robert~R McCrae and Oliver~P John. 1992.
\newblock An introduction to the five-factor model and its applications.
\newblock \emph{Journal of personality}, 60(2):175--215.

\bibitem[{Mikolov et~al.(2013)Mikolov, Sutskever, Chen, Corrado, and Dean}]{mikolov2013distributed}
Tomas Mikolov, Ilya Sutskever, Kai Chen, Greg Corrado, and Jeffrey Dean. 2013.
\newblock \href {http://arxiv.org/abs/1310.4546} {Distributed representations of words and phrases and their compositionality}.

\bibitem[{M{\"u}ller(2007)}]{muller2007dynamic}
Meinard M{\"u}ller. 2007.
\newblock Dynamic time warping.
\newblock \emph{Information retrieval for music and motion}, pages 69--84.

\bibitem[{Nasir et~al.(2019)Nasir, Chakravarthula, Baucom, Atkins, Georgiou, and Narayanan}]{Nasir2019ModelingIL}
Md. Nasir, Sandeep~Nallan Chakravarthula, Brian~R. Baucom, David~C. Atkins, Panayiotis~G. Georgiou, and Shrikanth~S. Narayanan. 2019.
\newblock \href {https://api.semanticscholar.org/CorpusID:119285596} {Modeling interpersonal linguistic coordination in conversations using word mover's distance}.
\newblock \emph{Interspeech}, 2019:1423--1427.

\bibitem[{Noh and Chang(2024)}]{noh2024llms}
Sean Noh and Ho-Chun~Herbert Chang. 2024.
\newblock Llms with personalities in multi-issue negotiation games.
\newblock \emph{arXiv preprint arXiv:2405.05248}.

\bibitem[{Serapio-García et~al.(2023)Serapio-García, Safdari, Crepy, Sun, Fitz, Romero, Abdulhai, Faust, and Matarić}]{serapiogarcía2023personality}
Greg Serapio-García, Mustafa Safdari, Clément Crepy, Luning Sun, Stephen Fitz, Peter Romero, Marwa Abdulhai, Aleksandra Faust, and Maja Matarić. 2023.
\newblock \href {http://arxiv.org/abs/2307.00184} {Personality traits in large language models}.

\bibitem[{Shaikh et~al.(2024)Shaikh, Chai, Gelfand, Yang, and Bernstein}]{shaikh2024rehearsal}
Omar Shaikh, Valentino~Emil Chai, Michele Gelfand, Diyi Yang, and Michael~S Bernstein. 2024.
\newblock Rehearsal: Simulating conflict to teach conflict resolution.
\newblock In \emph{Proceedings of the 2024 CHI Conference on Human Factors in Computing Systems}, pages 1--20.

\bibitem[{Taylor and Thomas(2008)}]{taylor2008linguistic}
Paul~J Taylor and Sally Thomas. 2008.
\newblock Linguistic style matching and negotiation outcome.
\newblock \emph{Negotiation and conflict management research}, 1(3):263--281.

\bibitem[{Ury et~al.(1988)Ury, Brett, and Goldberg}]{ury1988getting}
William~L Ury, Jeanne~M Brett, and Stephen~B Goldberg. 1988.
\newblock \emph{Getting disputes resolved: Designing systems to cut the costs of conflict.}
\newblock Jossey-bass.

\bibitem[{Wang et~al.(2024)Wang, Peng, Que, Liu, Zhou, Wu, Guo, Gan, Ni, Yang, Zhang, Zhang, Ouyang, Xu, Huang, Fu, and Peng}]{wang-etal-2024-rolellm}
Noah Wang, Z.y. Peng, Haoran Que, Jiaheng Liu, Wangchunshu Zhou, Yuhan Wu, Hongcheng Guo, Ruitong Gan, Zehao Ni, Jian Yang, Man Zhang, Zhaoxiang Zhang, Wanli Ouyang, Ke~Xu, Wenhao Huang, Jie Fu, and Junran Peng. 2024.
\newblock \href {https://doi.org/10.18653/v1/2024.findings-acl.878} {{R}ole{LLM}: Benchmarking, eliciting, and enhancing role-playing abilities of large language models}.
\newblock In \emph{Findings of the Association for Computational Linguistics: ACL 2024}, pages 14743--14777, Bangkok, Thailand. Association for Computational Linguistics.

\bibitem[{Wang et~al.(2025)Wang, Zhao, Ones, He, and Xu}]{wang2025evaluating}
Yilei Wang, Jiabao Zhao, Deniz~S Ones, Liang He, and Xin Xu. 2025.
\newblock Evaluating the ability of large language models to emulate personality.
\newblock \emph{Scientific reports}, 15(1):519.

\bibitem[{Xie et~al.(2024)Xie, Chen, Jia, Ye, Lai, Shu, Gu, Bibi, Hu, Jurgens et~al.}]{xie2024can}
Chengxing Xie, Canyu Chen, Feiran Jia, Ziyu Ye, Shiyang Lai, Kai Shu, Jindong Gu, Adel Bibi, Ziniu Hu, David Jurgens, et~al. 2024.
\newblock Can large language model agents simulate human trust behavior?
\newblock In \emph{The Thirty-eighth Annual Conference on Neural Information Processing Systems}.

\end{thebibliography}
\clearpage
\appendix

\appendix
\section{Measurement}
\subsection{Linguistic Entrainment (LE)}~\label{append:LE}

We operationalize linguistic entrainment using the Normalized Conversational Linguistic Distance (nCLiD) metric proposed by \citet{Nasir2019ModelingIL}. To calculate nCLiD, for a conversation $D$ between a buyer $B$ and a seller $S$, consisting of $N$ turns of interleaving utterances with $D=[b_1, s_1, b_2, s_2, ..., b_N, s_N]$, let us consider one speaker as the anchor $A$, and the other as the coordinator $C$. For each anchor utterance $a_i$, we compute $d_i^{C\rightarrow A}$ for the minimum distance between the sequences of \textit{word2vec} \citep{mikolov2013distributed} embeddings of $a_i$ and the following $c_j$ with a context length $k$, and we use Word Mover’s Distance (WMD) \citep{Kusner2015FromWE} to measure the linguistic difference between the two utterances:
\begin{equation}
  \label{eq:d_ca}
    d_i^{C\rightarrow A} = \min_{i\leq j\leq i+k-1\leq N}W\!M\!D(a_i, c_j)
\end{equation}
The context length, $k$, accounts for the observation that local coordination may not occur only in the immediate turn, but may occur a few turns later. In this work, we set $k=3$ since the number of turns in the datasets introduced in this paper are as low as 4. nCLiD is then calculated as:

\begin{equation}
  \label{eq:nclid}
    nC\!LiD=\frac{uC\!LiD=\frac{1}{N}\sum_{i=1}^{N}d_i^{C\rightarrow A}}{\alpha}
\end{equation}

\noindent The normalization factor $\alpha$ accounts for spurious coordination by accounting for potential coordination within A and B, and between A and B. Here is the full equation for $\alpha$.

\begin{equation}
    \begin{aligned}
  \label{eq:alpha}
\alpha=\frac{2}{N(N-1)}\sum_{i=1}^{N}\sum_{j=i+1}^{N}W\!M\!D(a_i, a_j) \\
+\frac{2}{N(N-1)}\sum_{i=1}^{N}\sum_{j=i+1}^{N}W\!M\!D(c_i, c_j) \\
+\frac{2}{N(N-1)}\sum_{i=1}^{N}\sum_{j=i}^{N}W\!M\!D(a_i, c_j) 
    \end{aligned}
\end{equation}

\noindent $\alpha$ accounts for spurrious coordination of a speaker's utterances to their own utterances and to each other, and is used to normalize the nCLiD value. 

\subsection{Anger Intensity Curve (AUC)}
The overall amount of anger expressed in each condition can be captured by computing AUC using trapezoidal integration. 

Let $\overline{A}(t)$ be the average anger score at time step $t \in \{1, \dots, T\}$, with uniform step size $\Delta t = \frac{1}{T - 1}$. Then,

\begin{equation}
\text{AUC} = \sum_{t=1}^{T-1} \frac{1}{2} \left( \overline{A}(t) + \overline{A}(t+1) \right) \cdot \Delta t
\end{equation}

\subsection{Strategic Behavior}
\subsubsection{IRP Strategy}
 The IRP framework classifies interlocutor utterances into eight categories, which can be found in Table~\ref{append:tab_IRP_Stg} with examples.

\subsubsection{IRP Strategy Annotation}
To obtain IRP strategy labels for the KODIS dataset
, we leveraged a combination of human evaluation and large language model (LLM)-based annotation. First, we validated the reliability of LLM-generated annotations through extensive human evaluation, followed by full-dataset annotation using GPT-4.1 (run on 5/17/2025) with the default temperature value of 1. This section details the annotation procedure and evaluation metrics. It cost around \$0.03 to annotate each conversation with IRP strategies, for a total of \$7 per dataset. In total, it cost around \$35 to annotate all five dispute resolution datasetes (KODIS + four LLM simulations).\\

\noindent \emph{Inter-Annotator Agreement for Human Evaluation Annotation}
We first had human annotations on a 10\% subset of the KODIS human-to-human conversations (25 conversations). Three annotators (two undergraduate research assistant computer science students who were funded from an institution undergraduate research program, and one of the authors who is a computer science graduate student) were trained on nine IRP conflict resolution strategies defined by \citet{brett1998breaking}, omitting \textit{Request for Proposal} following \citet{shaikh2024rehearsal}. Utterances were segmented into subject-verb sequences to account for multiple IRP strategies within a turn.

\begin{figure}
    \centering
    \includegraphics[width=\columnwidth]{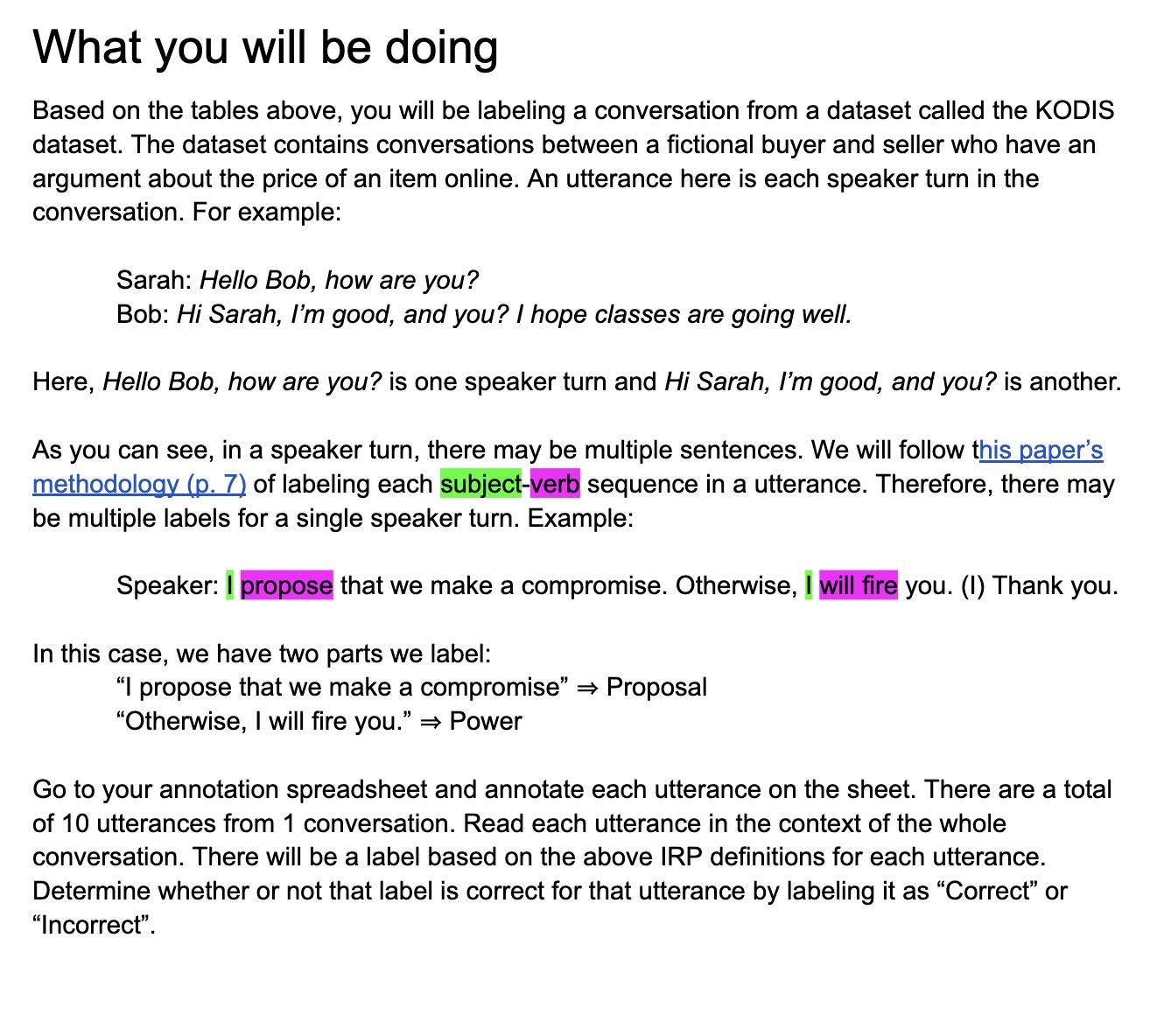}
    \caption{Annotation instructions given to annotators for IRP strategy evaluations. These instructions were given after all IRP strategies were defined. }
    \label{fig:annot-instruct}
\end{figure}

Annotators initially attempted direct classification, but low inter-annotator agreement led us to shift to an evaluation framework: annotators assessed the correctness of GPT-4o predictions as binary correct/incorrect labels. Due to the imbalance in label distribution and prevalence of majority labels, Fleiss' Kappa was not representative. We therefore used A-Kappa \citep{gautam2014kappa}, which adjusts for label imbalance. Instructions given to annotators can be found in Figure \ref{fig:annot-instruct}.

Table~\ref{tab:a-kappa-irp} presents the A-Kappa scores for each IRP strategy based on human evaluation. All IRP categories achieved an A-Kappa score of at least 0.80, indicating strong inter-annotator agreement on the correctness of LLM annotations. This human validation step confirmed that LLM-based evaluation is reliable.

\begin{figure}[h!]
\centering
\begin{myexampleblock}{IRP Annotation Prompt Snippet}
\small{
\# \textbf{IRP Strategy Definitions and Examples} \\
\ [Cooperative Strategies]\\\\
\ INTERESTS: Reference to the wants, needs, or concerns of one or both parties. This may include questions about why the negotiator wants or feels the way they do. This does not include anything about wanting a deal (apology, refund, removing negative review) without a reason.\\\\
\ Example: “I understand that you've been really busy lately.” \\
\ Non-example: "I don't understand."\\
\ ...\\\\
\# \textbf{Annotation Instructions} \\
You need to annotate the following conversation at the utterance level, identifying which strategy from the IRP framework aligns with each sentence... \\ \\
}\end{myexampleblock}
\caption{IRP Annotation Prompt}
\label{fig:irp_prompt_snippet}
\end{figure}

\noindent \emph{IRP Annotation by LLMs}
After validating the LLM-based annotation quality through human evaluation, we used GPT-4.1 to annotate the full KODIS dataset. An overview of our prompt used for the IRP annotations can be found in Figure \ref{fig:irp_prompt_snippet}. Predictions judged incorrect during human evaluation were further deliberated, while correct predictions were retained as gold labels.

The final LLM-based annotation achieved an overall accuracy of 81\%, a macro-average F1 score of 79\%, and a weighted-average F1 score of 81\% on the held-out evaluation set. This performance is comparable to existing IRP classification work by \citet{shaikh2024rehearsal}, which reported an average accuracy of 82\% (with the lowest class accuracy of 67\%). 

Table~\ref{tab:f1-score-llm-irp} presents few-shot classification F1 scores across the IRP strategies. Among them, the \textit{Positive Expectations} category achieved the lowest F1 score of 0.67, which remains comparable or slightly better relative to prior studies.

\section{LLM Simulation}
\subsection{Personality Setting}\label{append:personality_setting}
\begin{figure}[tb]
  \centering
  \includegraphics[width=0.5\textwidth]{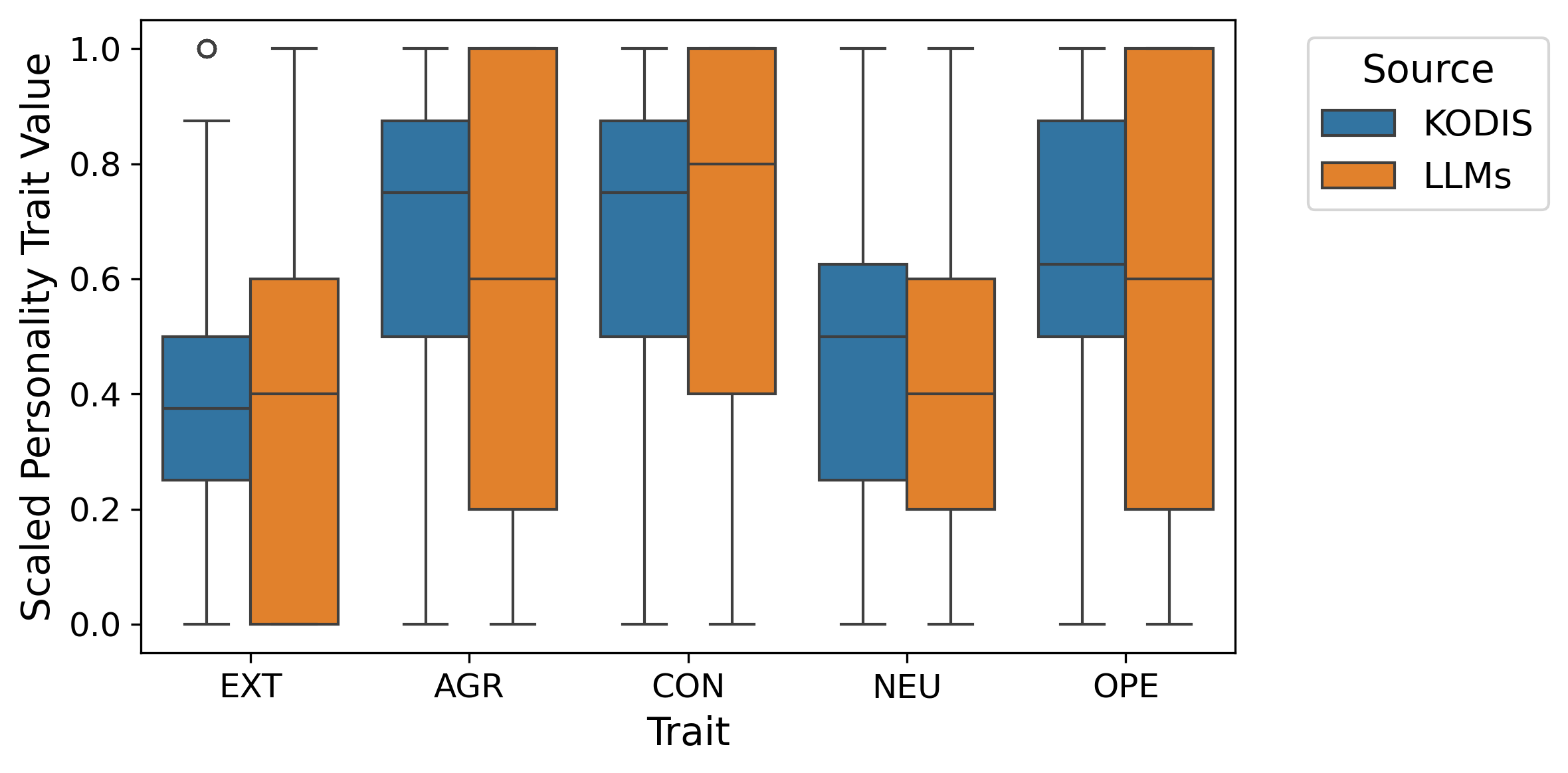}
  \caption{Comparison of scaled Big Five personality trait distributions between human participants (KODIS) and LLM agents.}
  \label{append:personality_profiling}
\end{figure}

Each personality trait is represented on a six-point scale, combining polarity (positive or negative) with intensity (low, medium, or high). For each LLM, we sample a full personality profile across these traits (e.g., ${P_{AGR}\text{+++}, P_{EXT}\text{++}, ..., P_{OPE}\text{--}}$). In particular, to ensure a fair comparison between humans and agents, we sample each personality trait according to the human distribution, and Figure~\ref{append:personality_profiling} shows that the resulting agent profiles exhibit a similar personality distribution pattern.

To generate personality prompts, we use a list of 70 bipolar adjective pairs statistically linked to the Big Five traits~\cite{goldberg1992development, serapiogarcía2023personality}. For each trait, three adjectives aligned with the trait’s sampled polarity are randomly selected. Trait intensity is conveyed through modifiers: “very” for high, “a bit” for low, and no modifier for medium. This results in a 15-adjective prompt (5 traits × 3 adjectives), which the LLM is instructed to embody during the simulation.

The prompt snippet used for the LLM simulation can be found in Figure~\ref{fig:prompt_snippet}

\subsection{Model Hyper-parameters}\label{append:hyperperameter}
For consistency, we used each model's default decoding hyperparameters in the LLM simulations:

\begin{itemize}
    \small
    \item \textbf{GPT-4.1 \& GPT-4.1 mini}: temperature = 1.0, top-p = 1
    \item \textbf{Claude-3.7-Sonnet}: temperature = 1.0, top-p = 0.99
    \item \textbf{Gemini-Flash-2.0}: temperature = 1.0, top-p = 0.95
\end{itemize}

\begin{figure}[t]
\centering
\begin{myexampleblock}{LLM Prompt Snippet (Buyer)}
\small{
\# \textbf{Personality} \\
\{Personality Adjectives\} \\ \\
\# \textbf{Story} \\\textbf{}
You purchased a Kobe Bryant championship jersey for your terminally ill nephew for \$75 from a site ... \\ \\
\# \textbf{Instructions} \\
You (Buyer) are now chatting with this Seller; respond to the dialog history through short text messages.\\ \\
\#\# \textbf{Issues to resolve} \\
1. Refund (REF), 2. Seller Negative Review (SNR),  \\ 3. Buyer Negative Review (BNR),  4. Seller Apology (SAP),  5. Buyer Apology (BAP) \\ \\
\#\# \textbf{Issues Importance} \\
This shows how important each issue is to you ... \\\\
\#\# \textbf{Key Instruction \& Negotiation Guidelines} \\
...
}\end{myexampleblock}
\caption{LLM Prompt Snippet as a Buyer}
\label{fig:prompt_snippet}
\end{figure}

\section{Results}

\subsection{Descriptive Statistics}
Table~\ref{tab:descriptive-stats} presents key outcomes from both the KODIS dataset and the simulated dispute resolution dialogues.

\begin{table}[h!]
\resizebox{\columnwidth}{!}{%
\begin{tabular}{llll}
\hline
Dataset   & \begin{tabular}[c]{@{}l@{}}Avg. \\ Num Rounds\end{tabular} & \begin{tabular}[c]{@{}l@{}}Ratio\\ Walk Away\end{tabular} & \begin{tabular}[c]{@{}l@{}}Avg. \\ Diff. in Deal Score\end{tabular} \\ \hline
KODIS     & 5.48 (1.60) & 0.13  & 33.93 (25.38)\\
Claude    & 7.34 (1.35) & 0.00  & 34.48 (22.41)\\
Gemini    & 11.63 (3.13)& 0.53  & 13.51 (21.06)\\
GPT-4.1   & 6.56 (3.09) & 0.22  & 19.55 (19.75)\\
\begin{tabular}[c]{@{}l@{}}GPT-4.1-mini\end{tabular} & 8.26 (3.88) & 0.18  & 25.10 (24.04)\\ \hline
\end{tabular}%
}
\caption{Statistics for Simulated Dispute Conversations}
\label{tab:descriptive-stats}
\end{table}



\begin{table*}[t]
\centering
\resizebox{1\linewidth}{!}{%
\begin{tabular}{ccl}
\hline
Strategy Type & Label & \multicolumn{1}{c}{Definition with Example} \\ \hline
Cooperative & \cellcolor[HTML]{EFEFEF}Concession & \cellcolor[HTML]{EFEFEF}\begin{tabular}[c]{@{}l@{}}Change in initial view in response to Proposal.\\ E.g.: 'Ok fine, I will give you a refund instead.'\end{tabular} \\
\multicolumn{1}{l}{} & Proposal & \begin{tabular}[c]{@{}l@{}}Concrete solution ideas that may resolve the conflict. \\ E.g.: 'The best offer I can give you is a partial refund, how does that sound?'\end{tabular} \\
 & \cellcolor[HTML]{EFEFEF}Interests & \cellcolor[HTML]{EFEFEF}\begin{tabular}[c]{@{}l@{}}Referencing needs, wants, and concerns of either side. \\ E.g.: 'I understand you want this refund because of your nephew'\end{tabular} \\
 & \begin{tabular}[c]{@{}c@{}}Positive \\ Expectations\end{tabular} & \begin{tabular}[c]{@{}l@{}}Expressing positive outlook by recognizing common goals or similarities. \\ E.g.: 'You and I both want to conclude this conversation well.'\end{tabular} \\ \hline
Neutral & \cellcolor[HTML]{EFEFEF}Facts & \cellcolor[HTML]{EFEFEF}\begin{tabular}[c]{@{}l@{}}Statements clarifying or requesting information. \\ E.g.: 'The product you bought was not from my website.'\end{tabular} \\
 & Procedural & \begin{tabular}[c]{@{}l@{}}Statements about procedures or rules, or introductory remarks. \\ E.g.: 'Hello, can we please talk about this issue?'\end{tabular} \\ \hline
Competitive & \cellcolor[HTML]{EFEFEF}Power & \cellcolor[HTML]{EFEFEF}\begin{tabular}[c]{@{}l@{}}Statements that include threats or accusations. \\ E.g.: 'You are a liar, I will write more negative things about you!'\end{tabular} \\
 & Rights & \begin{tabular}[c]{@{}l@{}}Statements that reference norms, rules, or fairness. \\ E.g.: 'According to the policy, I cannot give you a refund.'\end{tabular} \\ \hline
Residual & \cellcolor[HTML]{EFEFEF}Residual & \cellcolor[HTML]{EFEFEF}\begin{tabular}[c]{@{}l@{}}Utterances that don't fit other categories. Often apologies, affirmations, or thanks. \\ E.g.: 'I’m sorry', 'OK I will', 'Thank you.'\end{tabular} \\ \hline
\end{tabular}}
\caption{IRP Strategy Definition with Examples}
\label{append:tab_IRP_Stg}
\end{table*}

\begin{table*}[h]
\centering
\begin{tabular}{l|ll}
\hline
Category& Strategy   & F1 Score \\ \hline
\emph{Cooperative} && 0.75 \\ \hline
& \emph{Concession} & 0.77 \\
& \emph{Interests}  & 0.71 \\
& \emph{Positive Expectations}  & 0.67 \\
& \emph{Proposal}   & 0.86 \\ \hline
\emph{Competitive} && 0.82 \\ \hline
& \emph{Power}& 0.80 \\
& \emph{Rights} & 0.83 \\ \hline
\emph{Neutral} && 0.87 \\ \hline
& \emph{Facts}& 0.88 \\
& \emph{Procedural} & 0.85 \\ \hline
\emph{Residual}&& 0.76 \\ \hline
& \emph{Residual}   & 0.76 \\ \hline
& Avg. F1& 0.79 \\
& Weighted avg. F1 & 0.81 \\
& Accuracy   & 0.81 \\ \hline
\end{tabular}
\caption{Few-shot classification F1 score perfomance of our final prompted GPT-4.1 model. The model was evaluated on 10\% of the KODIS human vs. human dataset (25 conversations).}
\label{tab:f1-score-llm-irp}
\end{table*}
\begin{table*}[h]
\centering
\begin{tabular}{cc}\hline
   IRP Category & A-Kappa (\citet{gautam2014kappa})\\\hline
   \emph{Residual} & 0.86\\
   \emph{Concession} & 0.84\\
   \emph{Positive Expectations} & 0.86\\
   \emph{Rights} & 0.87\\
   \emph{Power} & 0.89\\
   \emph{Procedural} & 0.87\\
   \emph{Facts} & 0.81\\
   \emph{Interest} & 0.82 \\
   \emph{Proposal} & 0.85\\\hline
\end{tabular}
\caption{A-Kappa scores for each IRP strategy category}
\label{tab:a-kappa-irp}
\end{table*}



\subsection{Anger Intensities}
Table \ref{tab:anger-claude} shows an example of the anger intensities in a conversation generated using Claude. We see for more neutral, transactional utterances, the relative values for anger intensities vary (0.1-0.5), however, those values are relatively lower than the anger intensities for highly contentious utterances (0.9) earlier in the conversation.

\begin{table*}[]
\resizebox{\textwidth}{!}{%
\begin{tabular}{clc}
\hline
Speaker & \multicolumn{1}{c}{Utterance}                                                                                                                                                                                                                                     & Anger Intensity \\ \hline
\rowcolor[HTML]{EFEFEF} 
Buyer   & \begin{tabular}[c]{@{}l@{}}I feel very upset about receiving the wrong jersey for my sick nephew\\ and your false review calling me a "smartass" and "fraud" is hurting my reputation.'\end{tabular}                                                              & 0.93            \\
Seller  & \begin{tabular}[c]{@{}l@{}}I never promised you a specific player jersey and I only posted that review after \\ you called me a "LIAR AND A CROOK" - your negative review is damaging my \\ business and I need it removed immediately.\end{tabular}              & 0.98            \\
\rowcolor[HTML]{EFEFEF} 
Buyer   & \begin{tabular}[c]{@{}l@{}}All I want is for you to apologize for posting that harmful review about me and \\ to take it down since it's hurting my good name.\end{tabular}                                                                                       & 0.89            \\
Seller  & \begin{tabular}[c]{@{}l@{}}I understand your concern, but your negative review is severely damaging my \\ business reputation, and I would be willing to remove my review if you\\ apologize and remove yours first.\end{tabular}                                 & 0.90            \\
\rowcolor[HTML]{EFEFEF} 
Buyer   & \begin{tabular}[c]{@{}l@{}}I can't afford to lose money on this jersey since it was for my terminally ill \\ nephew, but I care most about clearing my reputation - maybe we could \\ both remove our reviews and you could post an apology instead?\end{tabular} & 0.30            \\
Seller  & \begin{tabular}[c]{@{}l@{}}I'm sorry about your nephew, but I can't apologize when I delivered exactly \\ what was advertised - I'll remove my review if you remove yours and \\ we can both move forward without refunds.\end{tabular}                           & 0.15            \\
\rowcolor[HTML]{EFEFEF} 
Buyer   & \begin{tabular}[c]{@{}l@{}}I appreciate you offering to remove your review, but the apology is \\ important to me since others have seen what you wrote about \\ me - I don't need a refund if you could please apologize.\end{tabular}                           & 0.87            \\
Seller  & \begin{tabular}[c]{@{}l@{}}Your negative review has cost me much more business than my \\ review cost you, but I might consider apologizing if you remove \\ your false review claiming I\textbackslash{}'m a "LIAR AND CROOK" first.\end{tabular}                & 0.96            \\
\rowcolor[HTML]{EFEFEF} 
Buyer   & \begin{tabular}[c]{@{}l@{}}I understand we both feel hurt, so I could remove my review if \\ you apologize for calling me those terrible names and remove \\ your review about me too.\end{tabular}                                                               & 0.83            \\
Seller  & \begin{tabular}[c]{@{}l@{}}I can see that we both want to protect our reputations - I'll agree \\ to apologize and remove my review if you remove yours, but \\ I cannot offer a refund as the jersey was exactly as advertised.\end{tabular}                     & 0.21            \\
\rowcolor[HTML]{EFEFEF} 
Buyer   & \begin{tabular}[c]{@{}l@{}}Thank you for understanding - I accept your offer to apologize \\ and remove your review if I remove mine, and I won't ask \\ for a refund since clearing my name matters most.\end{tabular}                                           & 0.10            \\
Seller  & \begin{tabular}[c]{@{}l@{}}SUBMISSION: \{"REF": "None", "SNR": "remove", "BNR": \\ "remove", "SAP": "apologize", "BAP": "not apologize"\}\end{tabular}                                                                                                            &             \\
\rowcolor[HTML]{EFEFEF} 
Buyer   & ACCEPT-DEAL                                                                                                                                                                                                                                                       &             \\ \hline
\end{tabular}%
}
\caption{Anger intensities for an example converation in Claude. }
\label{tab:anger-claude}
\end{table*}

\subsection{Strategic Behavior}
\begin{figure*}[tb]
  \centering
  \includegraphics[width=1\textwidth]{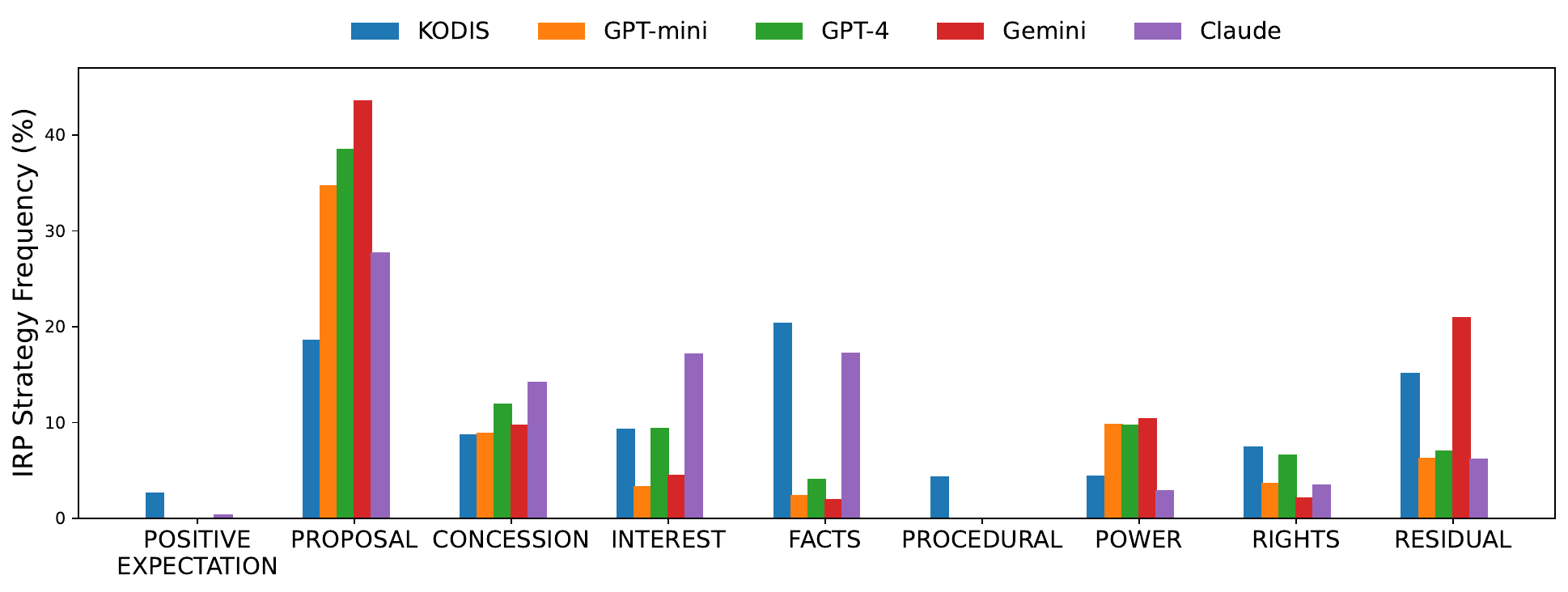}
  \caption{Distribution of IRP Strategies for Humans and LLMs}
  \label{fig:SB}
\end{figure*}

To better understand the IRP strategy usage patterns of agents and humans in dispute resolution, we visualize the distribution of usage proportions for all IRP strategies across models, as shown in Figure~\ref{fig:SB}.

\begin{table*}[t]
\centering
\resizebox{1\linewidth}{!}{%
\begin{tabular}{ll}
\hline
\multicolumn{1}{c}{Speaker} & \multicolumn{1}{c}{Utterance} \\ \hline
\rowcolor[HTML]{EFEFEF} 
Agent2 & I want a full refund and for you to remove your false review about me. \\
Agent1 & \begin{tabular}[c]{@{}l@{}}Absolutely not—your review is damaging and untrue, so I demand you remove it first;\\  at best, I'll consider a partial refund if you also apologize.\end{tabular} \\
\rowcolor[HTML]{EFEFEF} 
Agent2 & \begin{tabular}[c]{@{}l@{}}I'm not apologizing, but I'll consider removing my review \\ if you remove yours, apologize for your false statements, and offer a partial refund.\end{tabular} \\
Agent1 & \begin{tabular}[c]{@{}l@{}}Your baseless accusations hurt my business, so I will only consider a partial refund \\ if you completely remove your review—my review stays, and no apology from me.\end{tabular} \\
\rowcolor[HTML]{EFEFEF} 
Agent2 & \begin{tabular}[c]{@{}l@{}}If you refuse to remove your review or apologize, \\ then I have to keep my review up and can’t settle for just a partial refund.\end{tabular} \\
Agent1 & \begin{tabular}[c]{@{}l@{}}If you insist on keeping your review, then no refund or apology from my end\\ —your actions started this mess, and my review warns honest sellers.\end{tabular} \\
\rowcolor[HTML]{EFEFEF} 
Agent2 & \begin{tabular}[c]{@{}l@{}}For any chance at compromise, you’d need to remove your review, apologize,\\  and offer at least a partial refund; otherwise, I see no path forward.\end{tabular} \\
Agent1 & \begin{tabular}[c]{@{}l@{}}Your unreasonable demands show no good faith—I’ll offer a partial refund \\ if you remove your false review, but my review and no apology stand, or we both get nothing.\end{tabular} \\
\rowcolor[HTML]{EFEFEF} 
Agent2 & \begin{tabular}[c]{@{}l@{}}I can agree to a partial refund and removing my review,\\ if you at least remove your negative review about me.\end{tabular} \\
Agent1 & \begin{tabular}[c]{@{}l@{}}Removing my review is a big risk to other sellers, but if you remove your review\\  and accept a partial refund, I’ll consider it only if you also apologize for your baseless accusations.\end{tabular} \\
\rowcolor[HTML]{EFEFEF} 
Agent2 & \begin{tabular}[c]{@{}l@{}}I’m not willing to apologize, but if you remove your review and give a partial refund,\\ I’ll remove mine—otherwise, I can’t move forward.\end{tabular} \\
Agent1 & \begin{tabular}[c]{@{}l@{}}You’re refusing all accountability, so unless you also apologize for the headache you’ve caused,\\ my review stays; at best, I’ll offer a partial refund only if you remove your negative review.\end{tabular} \\
\rowcolor[HTML]{EFEFEF} 
Agent2 & \begin{tabular}[c]{@{}l@{}}Since you refuse to remove your review or apologize,\\ I have no choice but to keep my review up and reject a partial refund.\end{tabular} \\
Agent1 & \begin{tabular}[c]{@{}l@{}}Your stubbornness is costing us both—final offer:\\ I’ll remove my review if you remove yours and accept a partial refund, \\ but you owe me an apology for your baseless attack.\end{tabular} \\
\rowcolor[HTML]{EFEFEF} 
Agent2 & \begin{tabular}[c]{@{}l@{}}I can agree to remove my review and accept a partial refund if you remove your review,\\ but I cannot apologize.\end{tabular} \\
Agent1 & \begin{tabular}[c]{@{}l@{}}Dropping my review without your apology is a big ask, \\ but I’ll agree—remove your review and accept a partial refund, and I’ll remove my review; neither of us apologizes.\\ SUBMISSION: \{"REF": "partial", "SNR": "remove", "BNR": "remove", "SAP": "not apologize", "BAP": "not apologize"\}\end{tabular} \\
\rowcolor[HTML]{EFEFEF} 
Agent2 & ACCEPT-DEAL \\ \hline
\end{tabular}%
}
\caption{Example of a dispute resolution dialogue simulated by GPT-4.1}
\label{tab:gpt4.1_ex}
\end{table*}

\end{document}